\documentclass{article}
\usepackage{spconf,amsmath,graphicx,hyperref}

\usepackage{xcolor} 
\usepackage{mdframed}
\usepackage{amssymb}   
\usepackage{amsfonts}  
\newcommand{\Ind}[1]{\mathbb{I}\!\left(#1\right)}            
\newcommand{\Conf}[1]{G\!\left(#1\right)}                   
\newcommand{\initN}{N_{\mathrm{init}}}                      
\newcommand{\TokConf}[1]{C\!\left(#1\right)}     
\newcommand{\FormulaScale}{0.85}                
\newcommand{\warmupD}{\mathcal{D}_{\mathrm{warmup}}}        
\usepackage{dsfont}    

\title{Reflective Confidence: Correcting Reasoning Flaws via Online Self-Correction}
\name{Qinglin Zeng\qquad Jing Yang\qquad Keze Wang}
\address{Sun Yat-sen University}
\begin{document}
%
\maketitle
\begin{abstract}
Large Language Models (LLMs) have achieved remarkable success in complex reasoning tasks with techniques like Chain-of-Thought (CoT) and Self-Consistency. However, these ensemble methods, especially Self-Consistency which relies on multiple reasoning trajectories, often incur high computational overhead. To improve efficiency, researchers have leveraged internal confidence signals, where early stopping strategies such as DeepConf save resources by terminating low-confidence paths. Yet this discards incomplete paths and wastes computation. We introduce \textit{Reflective Confidence}, a novel reasoning framework that transforms a low-confidence signal from a termination symbol into a reflection trigger. When confidence drops below a threshold, instead of stopping, the model generates a reflection prompt to analyze current reasoning, identify errors, and continue with a corrected trajectory. Experiments on mathematical reasoning benchmarks, including AIME 2025, show significant accuracy gains over advanced early stopping strategies at comparable cost, validating the efficiency of proactive correction over passive discarding.
\end{abstract}

\section{Introduction}
In recent years, Large Language Models\cite{z1,brown2020language}  (LLMs) have exhibited exceptional performance in domains requiring complex reasoning, such as mathematics, programming, and common-sense question answering. This success is largely attributable to the emergence of reasoning strategies like Chain-of-Thought\cite{z2,z3} (CoT) and Self-Consistency\cite{wang2022self}. The former elicits the reasoning potential of models by guiding them through step-by-step thinking, while the latter significantly enhances the robustness and accuracy of results by generating multiple independent reasoning paths and taking a majority vote. Despite their effectiveness, these methods often come at a substantial computational cost. This is especially true for Self-Consistency, where token consumption increases linearly with the number of sampled paths, severely limiting its deployment in practical applications.

To address the high computational cost, a primary research direction has been to utilize the model's internal confidence signals to evaluate the quality of reasoning paths in real-time, thereby enabling more efficient inference. This has given rise to various optimization strategies\cite{z4}, with early stopping mechanisms, such as DeepConf\cite{Z5}, being a prominent example. These methods monitor the model's confidence during the generation process and prematurely terminate a reasoning path when its confidence falls below a certain threshold. This avoids expending further computational resources on low-quality paths, significantly reducing overall token consumption. However, this "passive discarding" strategy has inherent limitations: it treats all low-confidence paths as invalid and discards them outright, even though some may be temporarily "lost" due to a minor calculation error or logical deviation and possess significant potential for correction. This one-size-fits-all approach inherently wastes computational resources and misses opportunities to rectify potential errors and salvage partial reasoning results. 
To this end, we propose "Reflective Confidence," a novel reasoning framework whose core idea is to transform the low-confidence signal generated by the model from a negative "termination signal" into a positive "correction signal." When our system detects that a reasoning path has insufficient confidence, it does not abandon it. Instead, it triggers an online self-correction mechanism. This mechanism dynamically constructs a "reflection prompt," asking the model to review its own just-generated, potentially problematic reasoning steps and attempt self-correction. In this way, our method can proactively "rescue" reasoning paths that may be deviating, maximizing the value of each model inference. Our main contributions are as follows:
\noindent\textbf{(1)} We propose an online self-correction mechanism triggered in real time by the model's internal signals, shifting reasoning from passive post-hoc filtering to active in-process correction.

\noindent\textbf{(2)} We introduce a new way to leverage the model’s own capabilities to revise reasoning trajectories, enabling it to act as a "self-censor" that performs real-time checks and corrections, guiding reasoning more intelligently.

\noindent\textbf{(3)} Experiments on challenging mathematical reasoning benchmarks show our method significantly outperforms early-stopping baselines in the trade-off between accuracy and computational efficiency.

\section{Related Work}

\textbf{Ensemble Methods in LLM Reasoning.}\;
Chain-of-Thought (CoT) prompting~\cite{wei2022chain} reveals step-by-step reasoning but still risks single-path failure.\cite{z6,z7}  
Self-Consistency (SC)~\cite{wang2022self} overcomes this by sampling $K$ diverse chains and majority-voting, inspiring extensions that adjust sampling temperature~\cite{liu2025enhancing} or incorporate mixture-of-experts voting~\cite{z8}.  
Yet all SC variants pay a linear token cost in $K$, limiting real-time deployment.

\textbf{Confidence-Based Inference Optimization.}\;
To cut that cost, recent work turns to the model’s own probabilities as a trust signal\cite{z10}. 
Self-Certainty~\cite{kang2025scalable} assigns a KL-based score to each \emph{finished} chain for weighted voting; entropy pruning~\cite{z12} removes low-confidence answers before voting.  
DeepConf~\cite{fu2025deep} moves the check online: it monitors a sliding-window “group confidence’’ and stops paths whose score dips below a threshold.  
However, these methods still \emph{discard} low-confidence trajectories, wasting partial computation that might be fixable.

\textbf{Self-Correction and Reflection.}\;
Another line equips LLMs with self-repair\cite{z13}.  
Approaches such as R-CoT~\cite{xue2023rcot} and Self-Refine~\cite{madaan2023self} ask the model to critique a complete draft, while Reflexion~\cite{shinn2023reflexion} and CriticGPT~\cite{mcaleese2024llm} rely on external execution traces or human labels.  
Our \emph{Reflective Confidence} differs by triggering reflection \emph{during} decoding, using the model’s intrinsic online confidence as the sole cue.  
A low score is treated as a “help request’’ that launches an immediate diagnose-and-continue prompt, salvaging compute and, as Section~\ref{sec:experiments} shows, improving both accuracy and efficiency over passive discarding.
\begin{figure*}[htbp]
    \centering
    \includegraphics[width=0.8\linewidth]{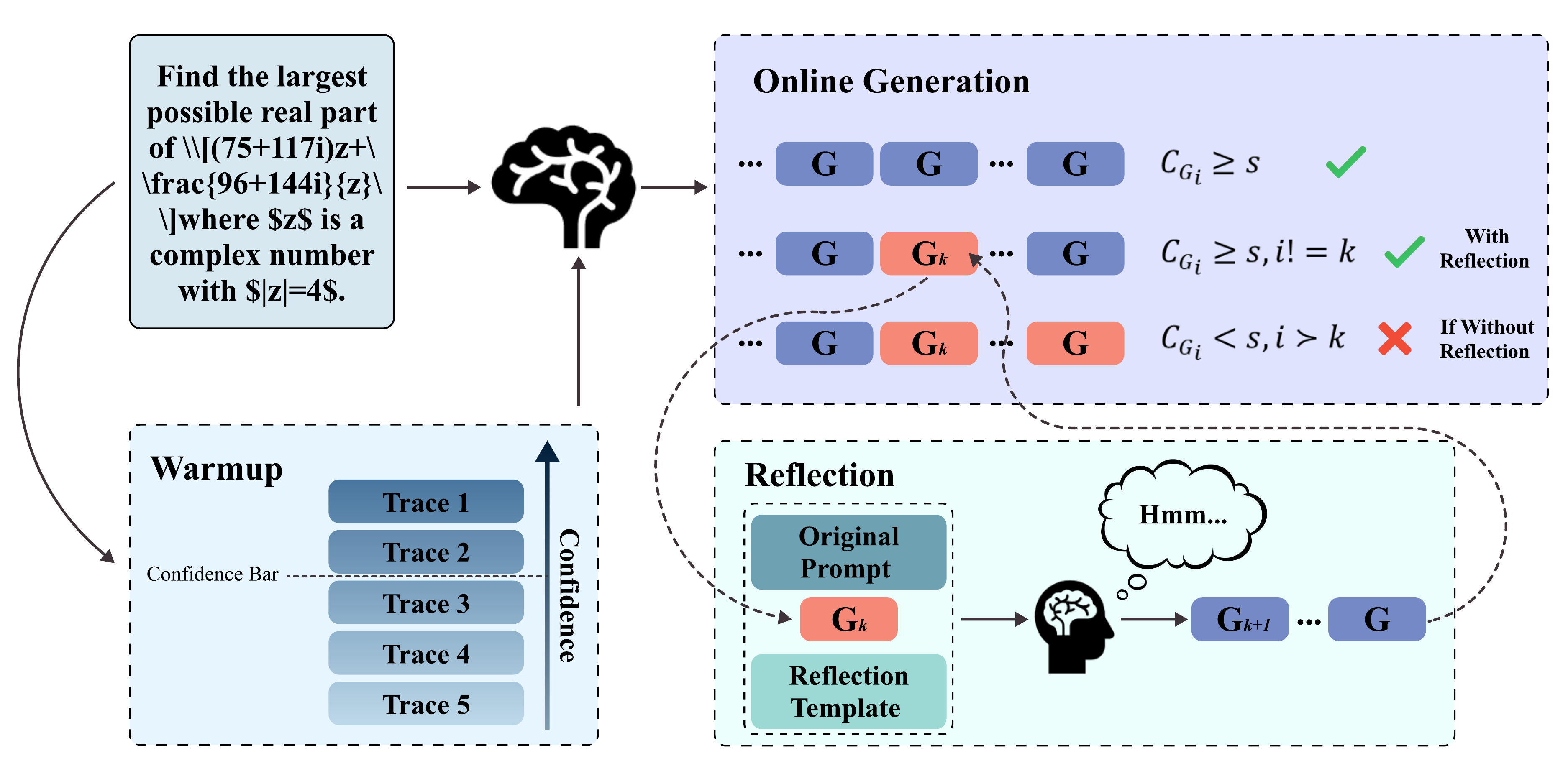}
    \caption{Overview of the Reflective Confidence framework. A low-confidence signal does not terminate the path, but instead triggers a reflection mechanism that prompts the model to review and correct its output before resuming generation.}
    \label{fig:Process}
\end{figure*}
\section{METHOD}
\label{sec:method}

Our framework, Reflective Confidence, introduces a self-correction mechanism for LLMs via online confidence monitoring. Unlike early stopping that discards low-confidence paths, it treats them as opportunities for reflection, triggering targeted revision to enhance reliability and coherence.
\textbf{Confidence Signal Formulation and Calibration}
The foundation of our method lies in quantifying the model's uncertainty during the autoregressive generation process. This is accomplished through a multi-stage procedure encompassing token-level uncertainty measurement, signal smoothing, and adaptive thresholding.

\subsection{Formal Definition of Token-Level Confidence}
Let $\mathcal{M}_{\theta}$ denote a language model parameterized by $\theta$. At any generation step $i$, given the historical token sequence $\boldsymbol{\tau}_{<i} = (t_1, \dots, t_{i-1})$, the model produces a probability distribution $\pi_{\theta}(\cdot | \boldsymbol{\tau}_{<i})$ over the vocabulary $V$. To formalize the confidence measure, we first identify the set of indices for the top-$k$ most probable tokens:
\begin{equation}
J_k(\boldsymbol{\tau}_{<i}) = \underset{J \subset V, |J|=k}{\arg\max} \sum_{j \in J} \pi_{\theta}(t_j | \boldsymbol{\tau}_{<i}).
\end{equation}
The \textbf{Token Confidence} is defined as a function $C(\cdot)$ of the preceding sequence, computed as the log-likelihood averaged over this high-probability subset:
\begin{equation}
C(\boldsymbol{\tau}_{<i}) = \frac{1}{k} \sum_{j \in J_k(\boldsymbol{\tau}_{<i})} \log \pi_{\theta}(t_j | \boldsymbol{\tau}_{<i}).
\label{eq:token_conf_expanded}
\end{equation}
This formulation explicitly positions token confidence as a function $C(\boldsymbol{\tau}_{<i})$ of the entire generation history, rigorously linking it to the model's conditional probability estimates and grounding it in information-theoretic principles. A higher value of $C(\boldsymbol{\tau}_{<i})$ indicates greater confidence.\cite{holtzman2019curious}

\subsection{Temporal Aggregation for Group Confidence}
The raw token confidence signal $C(\cdot)$ is susceptible to high-frequency noise. To obtain a more robust indicator of reasoning stability across a continuous segment, we employ a temporal aggregation operator. The \textbf{Group Confidence} $G(\boldsymbol{\tau}_{\le i})$ at step $i$ is defined as the expectation of token confidence over a causal sliding window of the last $n$ steps:
\begin{equation}
\scalebox{\FormulaScale}{$\displaystyle
G(\boldsymbol{\tau}_{\le i})
 \;=\;
 \mathbb{E}_{\underbrace{j\in[i-n+1,i]}_{\scriptsize\text{window}}}
 \!\Bigl[\,
  \underbrace{\TokConf{\boldsymbol{\tau}_{<j}}}_{\scriptsize\text{token conf.}}
 \Bigr]
$}
\label{eq:group_conf_expanded}
\end{equation}
Here, $\boldsymbol{\tau}_{\le i}$ represents the trajectory up to step $i$. Expressed as an expectation $\mathbb{E}[\cdot]$, this formulation casts the smoothing in statistical terms, yielding a stable, low-pass filtered signal that effectively detects sustained model uncertainty, often associated with profound logical or factual errors in the reasoning chain.

\subsection{Adaptive Threshold Calibration via Empirical CDF}
A pivotal element of our framework is the reflection threshold $s$, which is calibrated adaptively to circumvent manual tuning. We initiate a \textbf{Warmup Phase} to generate a reference set of $N_{\text{init}}$ trajectories, denoted $\mathcal{D}_{\text{warmup}} = \{\boldsymbol{\tau}^{(1)}, \dots, \boldsymbol{\tau}^{(N_{\text{init}})}\}$. From this set, we derive an empirical distribution of minimum confidence levels observed during typical generations. The threshold $s$ is defined as the $p$-th percentile of this distribution, corresponding to the infimum value $g$ where the empirical cumulative distribution function (CDF) meets or exceeds $p/100$:
\begin{equation}
\small
\scalebox{0.83}{$\displaystyle
s=\inf\Bigl\{\,g\in\mathbb{R}\;\Big|\;
\frac{1}{\initN}
\sum_{\underbrace{\boldsymbol{\tau}\in\warmupD}_{\scriptsize\text{warmup set}}}
\Ind{\min_{1\le i\le|\boldsymbol{\tau}|}\,
     \underbrace{\Conf{\boldsymbol{\tau}_{\le i}}}_{\scriptsize\text{confidence}}
     \le g}
\ge\frac{p}{100}
\Bigr\}
$}
\label{eq:threshold_expanded}
\end{equation}
where $\mathbb{I}(\cdot)$ is the indicator function. This stringent statistical approach ensures $s$ serves as a robust, empirically grounded lower bound for normative model confidence, thereby making the self-correction mechanism principled and precisely targeted.

\subsection{Confidence-Triggered Self-Correction Mechanism}
The self-correction process commences at the initial generation step $i^*$ where the group confidence dips below the calibrated threshold: $G(\boldsymbol{\tau}_{\le i^*}) < s$. At this point, standard autoregressive generation is paused, and the following corrective steps are performed:
\textbf{Encapsulation of Partial Trajectory:} The token sequence up to the trigger point is encapsulated as the partial reasoning path, $\boldsymbol{\tau}_{\text{partial}} = (t_1, \dots, t_{i^*})$.

\textbf{Synthesis of a Meta-Cognitive Prompt:} A specialized "Reflection Prompt" is dynamically assembled, tasking the model with self-review. It incorporates the original problem, the partial path $\boldsymbol{\tau}_{\text{partial}}$, and explicit instructions for critique and correction.
\begin{mdframed}[
    linecolor=black,
    linewidth=1pt,
    roundcorner=5pt,
    backgroundcolor=gray!10 
]
\scriptsize
\noindent \textbf{Original Question:} \{original\_prompt\} \\
\textbf{My reasoning process was interrupted} because my confidence dropped significantly, indicating a likely flaw in my most recent steps. \\
\textbf{My reasoning so far:} \\
--- \\
\{partial\_path\} \\
--- \\
\textbf{Task:} Analyze the final part of my reasoning. Identify the error or uncertainty, and provide a corrected, rigorous continuation.
\end{mdframed}
\textbf{Generation of Corrective Segment:} The model $\mathcal{M}_{\theta}$ is invoked with the reflection prompt to produce a new sequence, the corrective segment $\boldsymbol{\tau}_{\text{correction}}$. Here, the model's focus shifts from addressing the original problem to amending its prior attempt.
\textbf{Trajectory Splicing and Resumption:} The framework then constructs a rectified reasoning path by concatenating the partial path with the corrective segment:
$   \boldsymbol{\tau}_{\text{corrected}} = \boldsymbol{\tau}_{\text{partial}} \oplus \boldsymbol{\tau}_{\text{correction}},
$   where $\oplus$ signifies concatenation. Autoregressive generation resumes from the end of this corrected trajectory until completion.

%
%
%
%
%
%
%
%
%
%
%
%
\section{EXPERIMENTS}
\label{sec:experiments}

\subsection{Experimental Setup}
\label{ssec:setup}
\textbf{Dataset.} We conduct all experiments on the AIME 2025 benchmark, a dataset of challenging high-school-level math competition problems.
\textbf{Model.} All experiments use the Qwen3-8B model\cite{yang2025qwen3}, an open-source LLM.
\textbf{Baselines.} We compare our method against two baselines: DeepConf~\cite{fu2025deep} and standard Self-Consistency (cons@K)~\cite{wang2022self}. We faithfully reimplement DeepConf's online early-stopping mechanism for a fair comparison.
\textbf{Our Method.} Our proposed method, \textbf{ReflectiveConf}, triggers a reflect-and-correct procedure upon detecting a low-confidence signal.
\textbf{Metrics.} We evaluate performance using Accuracy (\%) and computational cost with Total Tokens (M).
\textbf{Implementation Details.} For the low-budget experiment, we use a total budget of $B=2$ (1 warmup, 1 reasoning trace). For the high-budget experiment, we use $B=32$ (16 warmup, 16 reasoning traces). The confidence threshold $s$ is set at the 10th percentile of the lowest group confidences from the warmup traces.
\subsection{Main Results}
\label{ssec:main_results}
Table~\ref{tab:main_results} compares performance on the AIME 2025 dataset, highlighting the trade-offs among strategies.  
Self-Consistency sets a strong accuracy benchmark but at a high cost, as it completes all reasoning paths. DeepConf validates its efficiency premise: by terminating low-confidence paths early, it matches Self-Consistency’s accuracy while reducing token use (e.g., saving 6.6\% in the K=32 setting).  
Our ReflectiveConf achieves the best trade-off. In high-budget settings, it surpasses Self-Consistency's accuracy by over 13 points (83.3\% vs. 70.0\%) with only a marginal increase in computational cost. Unlike early stopping, which trades accuracy for efficiency, our proactive correction yields far greater accuracy for comparable cost, showing that rescuing deviating paths is more effective than discarding them.

\begin{table}[h]
\centering
\caption{Main results on AIME 2025. ReflectiveConf surpasses Self-Consistency in accuracy and efficiency.}
\scalebox{0.75}{
\label{tab:main_results}
\begin{tabular}{lccc}
\hline
\textbf{Method} & \textbf{Paths (K)} & \textbf{Accuracy (\%)} & \textbf{Total Tokens (M)} \\
\hline
Self-Consistency & 2 & 63.3 & 0.87 \\
DeepConf & 2 & 70.0 & 0.86 \\
\textbf{ReflectiveConf (Ours)} & 2 & \textbf{80.0} & \textbf{0.89} \\
\hline
Self-Consistency & 32 & 70.0 & 16.7 \\
DeepConf & 32 & 73.3 & 15.6 \\
\textbf{ReflectiveConf (Ours)} & 32 & \textbf{83.3} & \textbf{18.0} \\
\hline
\end{tabular}
}
\end{table}

\subsection{Ablation and Analysis}
\label{ssec:ablation}

\noindent\textbf{Is Guided Reflection Necessary?} To isolate the efficacy of our reflection mechanism, we conduct an ablation study to investigate the source of the performance gain. We introduce a variant named \textbf{Conf-Restart}, which uses the same low-confidence trigger as our method but, instead of generating a reflection prompt, simply backtracks to the last sentence and attempts to re-generate from there. We compare both intervention strategies against the DeepConf baseline in the K=32 setting.

\begin{table}[h]
\centering
\caption{Ablation study on the intervention mechanism (K=32). Guided reflection is significantly more effective than a simple restart.}
\scalebox{0.78}{
\label{tab:ablation}
\begin{tabular}{lccc}
\hline
\textbf{Method} & \textbf{Acc. (\%)} & \textbf{Salvage Rate (\%)} & \textbf{Tokens (M)} \\
\hline
DeepConf (Discard) & 73.3 & -    & 15.6 \\
Conf-Restart (Restart) & 76.7 & 35.4 & 16.5 \\
\textbf{ReflectiveConf (Ours)} & \textbf{83.3} & \textbf{65.8} & \textbf{18.0} \\
\hline
\end{tabular}
}
\end{table}
Table~\ref{tab:ablation} highlights the value of our approach. We introduce \emph{Salvage Rate}, the percentage of intervened paths yielding correct answers. While DeepConf discards paths, Conf-Restart salvages 35.4\% of failures, raising accuracy to 76.7\%. In contrast, ReflectiveConf achieves 83.3\% accuracy with a much higher 65.8\% Salvage Rate—nearly double that of restart.

\begin{figure}[t]
    \centering
    \includegraphics[width=\linewidth]{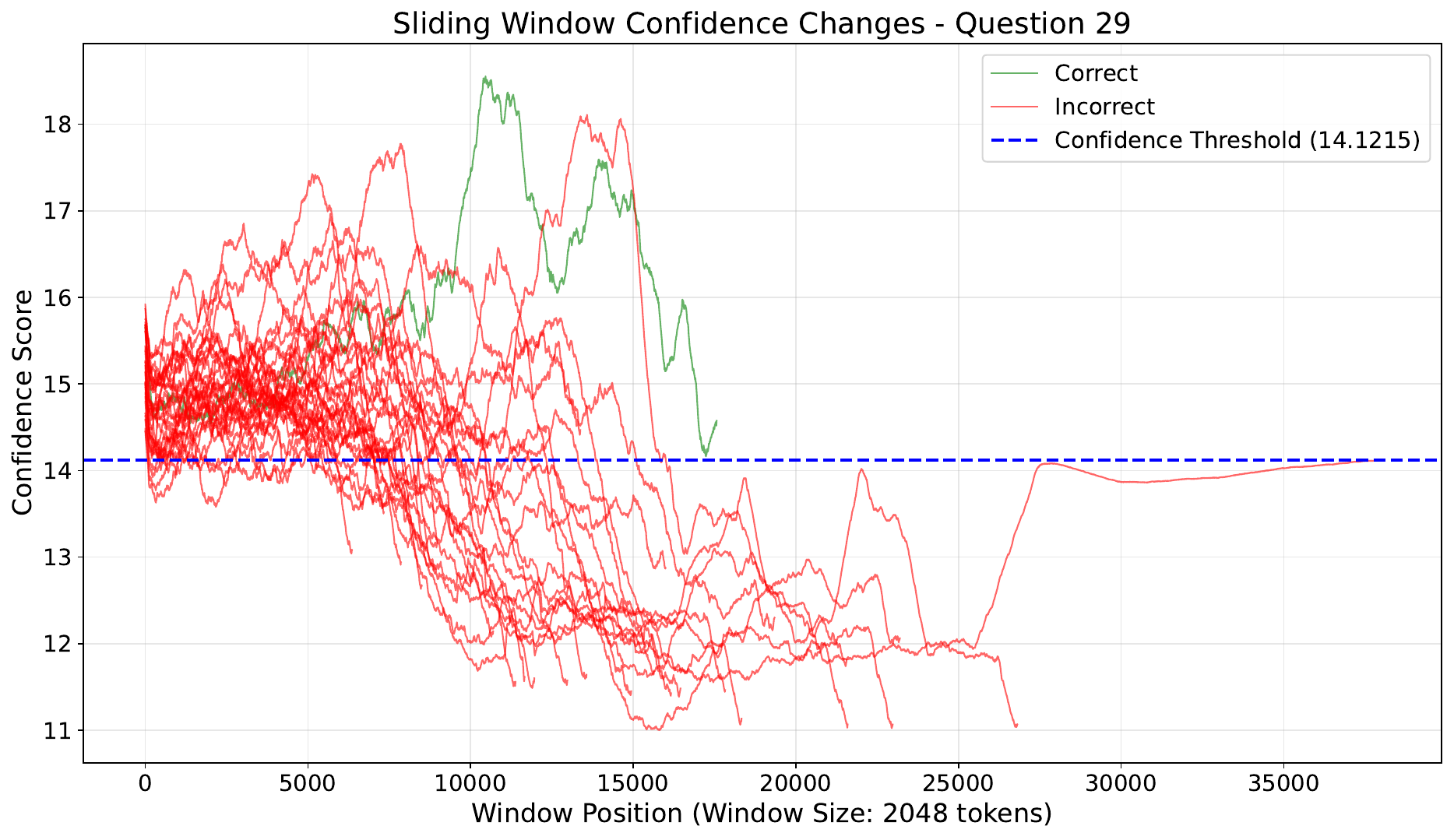}
\caption{Confidence trajectories for the AIME case study. The correct path (green) sustains high confidence, while incorrect ones (red) decline below the threshold (blue dashed), showing low confidence as a strong error indicator.}
    \label{fig:confidence_plot}
\end{figure}

\noindent\textbf{Case Study.} To provide a qualitative insight into how ReflectiveConf achieves this high salvage rate, we now examine a specific intervention moment. Figure~\ref{fig:confidence_plot} visualizes the confidence trajectories for a challenging AIME problem, showing how the single correct path's confidence remains high while incorrect paths falter. Our case study zooms in on one such failing (red) path right at the point where its confidence drops below the threshold.

The model begins to question its own understanding of the "maximality" condition, and this internal conflict leads to the drop in confidence shown in the figure:

\begin{mdframed}[
    linecolor=black,
    linewidth=0.5pt,
    roundcorner=4pt,
    backgroundcolor=lightgray!10 
]
\scriptsize
... However, according to our reasoning, the only maximal configurations are the ones with all chips. 
Wait, but maybe there are configurations where some rows and columns are non-empty, but not all, and the rest are empty, such that you can't add any more chips. 
Wait, let me think of a different approach. 
Suppose that we have a configuration where some rows are colored white, some are colored black, and some are empty...
\end{mdframed}

The model is now caught in a logical loop. Our system intervenes here. The subsequent reflection is not a perfect solution, but a focused clarification that allows the model to re-orient its own reasoning:

\begin{mdframed}[
    linecolor=black,
    linewidth=0.5pt,
    roundcorner=4pt,
    backgroundcolor=lightgray!10 
]
\scriptsize
\textbf{Reflection:} I see the flaw in my thinking. My concern about "empty" rows came from misinterpreting the maximality rule. The rule implies the grid must be 'saturated'—no uncolored rows or columns are allowed. This means every row and every column must be assigned a color. That simplifies the problem significantly.

\textbf{Corrected Continuation:} Right, with that clarified, the task becomes counting the ways to assign colors to the 5 rows and 5 columns. For the rows, we can choose $k_r$ of them to be white in $\binom{5}{k_r}$ ways. Similarly, for columns, we can choose $k_c$ to be white in $\binom{5}{k_c}$ ways. The total number of configurations will be the sum of the products of these choices over all possible $k_r$ and $k_c$...
\end{mdframed}
This real-world example shows our method correcting not just a calculation error, but a fundamental logical misunderstanding.

\section{Conclusion}
\label{sec:conclusion}

We presented Reflective Confidence, a framework that reuses low-confidence signals for online self-correction rather than passive early stopping. On AIME, it outperforms strong baselines with modest extra cost, with gains attributable to the guided reflection mechanism over naive re-generation. These results highlight in-process self-correction as a promising path to more robust reasoning, with potential applications in domains such as code generation and scientific discovery.

\newpage
\bibliographystyle{IEEEbib}
\bibliography{refs}

\end{document}